\documentclass[letterpaper]{article}

\usepackage{times}
\usepackage{helvet}
\usepackage{courier}
\usepackage{subfig}
\usepackage{url}
\usepackage[pdftex]{graphicx}
\usepackage[utf8]{inputenc}
\usepackage[T1]{fontenc}
\usepackage{booktabs}
\usepackage{amsfonts}
\usepackage{nicefrac}
\usepackage{microtype}
\usepackage{tikz}
\usepackage{grffile}
\usetikzlibrary{shapes.misc, positioning}
\usepackage{thmtools,thm-restate}
\usepackage{parskip}
\usepackage{authblk}
\usepackage{tcolorbox}
\usepackage{wrapfig}
\usepackage{amsmath}
\usepackage{amssymb}
\usepackage{mathtools}
\usepackage{amsthm}
\usepackage[numbers]{natbib}
\usepackage{multirow}
\usepackage{color}
\usepackage{hyperref}
\usepackage{colortbl}
\usepackage{algpseudocode}
\usepackage[none]{hyphenat}
\usepackage{floatrow}
\usepackage{macros}
\usepackage[pro]{fontawesome5}
\usepackage{basic}

\hypersetup{
    colorlinks=true,
    linkcolor=purple,
    urlcolor=teal,
    linktoc=all,
    citecolor=teal
}

\newfloatcommand{capbtabbox}{table}[][\FBwidth]

\title{ScriptoriumWS: A Code Generation Assistant for Weak Supervision}

\renewcommand{\algorithmicrequire}{\textbf{Input:}}
\renewcommand{\algorithmicensure}{\textbf{Output:}}

\author{%
  Tzu-Heng~Huang, Catherine~Cao, Spencer~Schoenberg, Harit~Vishwakarma, Nicholas~Roberts, Frederic Sala\\
  University of Wisconsin-Madison \\
  \texttt{\{thuang273, ccao35, spencer.schoenberg\}@wisc.edu}, \\
  \texttt{\{hvishwakarma, nick11roberts, fredsala\}@cs.wisc.edu} \\
}

\begin{document}
\date{}

\maketitle
\begin{abstract}
Weak supervision is a popular framework for overcoming the labeled data bottleneck: the need to obtain labels for training data. 
In weak supervision, multiple noisy-but-cheap sources are used to provide guesses of the label and are aggregated to produce high-quality pseudolabels.
These sources are often expressed as small programs written by domain experts---and so are expensive to obtain.
Instead, we argue for using code-generation models to act as coding assistants for crafting weak supervision sources.
We study prompting strategies to maximize the quality of the generated sources, settling on a multi-tier strategy that incorporates multiple types of information.
We explore how to best combine hand-written and generated sources.
Using these insights, we introduce ScriptoriumWS, a weak supervision system that, when compared to hand-crafted sources, maintains accuracy and greatly improves coverage.
We release our code and results here: \url{https://github.com/SprocketLab/ScriptoriumWS}~\footnote{Appeared in ICLR'23 Deep Learning for Code (DL4C) Workshop \& 2023 Midwest Machine Learning Symposium.}.
\end{abstract}

\section{Introduction}

%% What is the problem?
%% introduce PWS
%% label functions in PWS are hard to design
%% create label function manually is time-consuming and costly.

Access to substantial amounts of high-quality labeled data is a key ingredient for training performant machine learning models. Such data is usually produced by asking domain experts for ground-truth labels, making the process of dataset creation expensive, slow, and hard to scale. Programmatic weak supervision (PWS), a novel paradigm for generating labeled data \cite{ratner2016data}, sidesteps these obstacles. The idea behind PWS is to leverage a combination of noisy label estimates obtained from domain knowledge, heuristic rules, and pattern matching. These sources act as noisy labeling functions (LFs), usually expressed as \textbf{code}. The outputs of these labeling functions are modeled and aggregated to annotate unlabeled data points \cite{ratner2016data, ratner2017snorkel, ratner2019training, fu2020fast}. 

PWS has proven successful \cite{bach2019snorkel, evensen2020ruler, li2021weakly, gao2022classifying} but remains expensive: users must painstakingly write small programs to act as LFs. Users, even domain experts, often need tedious experimentation to carefully set up proper thresholds, manually fine-tune heuristic rules to capture enough keywords, or debug regular expressions. %It is also  challenging for those without domain knowledge to discover sufficient insights and helpful rules to design LFs for specific scenarios.
%
%In recent years, to reduce the effort required in handcrafting LFs, 
To tackle these challenges, recent approaches automatically produce %, which aim at automating the 
LFs by using a minimal level of supervision (i.e. a few labeled data points) \cite{varma2018snuba, das2020goggles, inproceedings, boecking2021interactive, roberts2022autowsbench} or access to powerful external models (like large language models) to prompt data labels \cite{smith2022language}. However, these approaches do not yield programmatic LFs, but rather model-generated noisy label estimates, and so lose the ability to debug and transfer, a key advantage of programmatic weak supervision.

A best-of-both worlds approach is to have \textbf{code-generation models write labeling functions}. This neither requires domain experts to write code nor sacrifices the programmatic property of LFs. 
Indeed, such an approach is now plausible given advances in models that produce code, such as CodeT5 \cite{wang2021codet5}, Codex \cite{chen2021evaluating}, and CodeGen \cite{Nijkamp2022CG}). 
Among other benefits, LFs generated by such models can be edited and used as templates, providing programming assistance for users to design LFs more easily and efficiently. 
Additionally, unlike human-designed LFs, synthesized LFs can be generated in large quantities. 
Finally, in contrast to using large language models to obtain the noisy labels estimates via prompting, which requires repeated inference calls, synthesized LFs can be stored and reused to label new data at zero cost. %With synthesized LFs, users do not need domain expertise to design LFs. Writing prompts becomes much easier for users to leverage rich knowledge pre-trained on code into the LF generation.

%% Why is it hard? (E.g., why do naive approaches fail?)
%% prompt is sensitive, hard to control the quality of LFs, no one knows the trade-off, what's the correct way to leverage LFs into PWS
%% Codex is not open-source, leaving many questions to design the correct format to compose a prompt to generate high-quality label functions.
%% it is still unclear about the proper way to bring benefits of code LLMs into data labeling. 

However, it is unclear whether code generation models can produce sufficiently high-quality LFs, and, when it is possible, what approach to take in order to do so. We ask the following fundamental questions we aim to answer in this work:
\begin{enumerate}
    \item \textbf{Prompt format}: Prompts are highly sensitive. Small changes to prompt components lead to great variation in generated results. There is currently no consensus on the best way to generally prompt code-generation models, let alone specifically for labeling functions. Our first question is: what prompting strategy can yield high-quality LFs?
    \item \textbf{Capability of synthesized LFs}: Next we ask: compared to human-designed LFs, what are the strengths and weaknesses of synthesized LFs? Additionally, what is the typical result when using these synthesized programs in programmatic weak supervision pipelines?
    \item \textbf{In-context few-shot settings}: If we are allowed to include some heuristic rules or give several data examples into the prompt context, does this better guide the model in synthesizing high-quality LFs? What type of in-context information can add and influence the quality of synthesized LFs?
\end{enumerate}

%% Why is it interesting and important?
%% Foundation models can be powerful knowledge source

\begin{figure}[t!]
    \includegraphics[width=\linewidth]{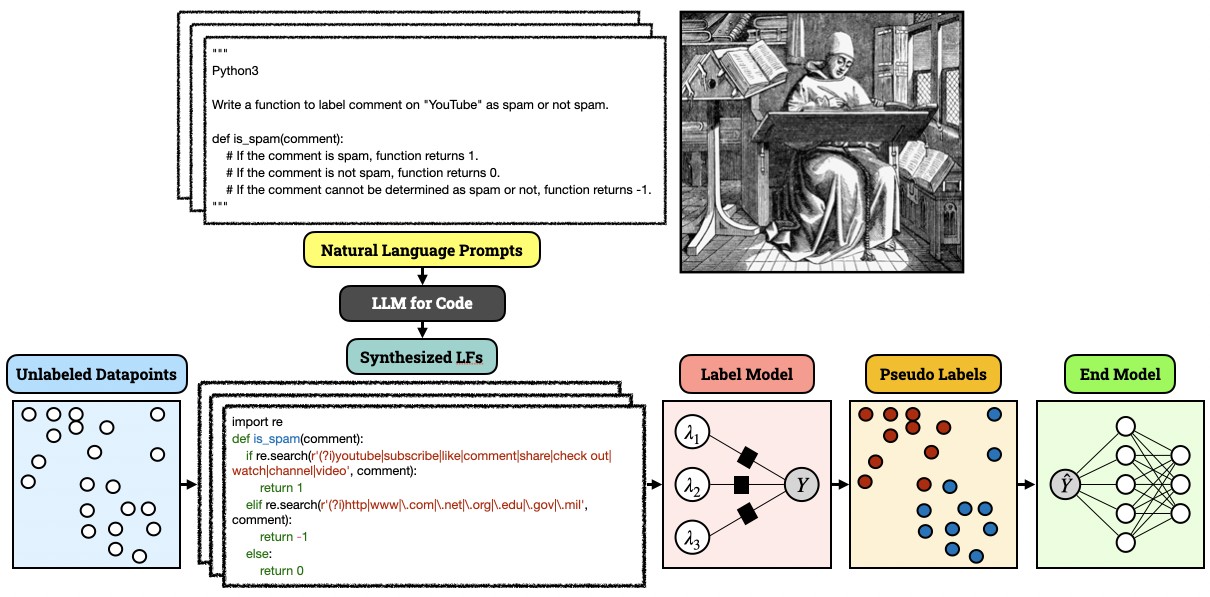}
    \centering
    \caption{\small Overview of the proposed ScriptoriumWS system. Code generation models are prompted to produce small programs that act as weak supervision labeling functions. These are used within a weak supervision pipeline to label an unlabeled dataset. A downstream end model is trained on the labeled data.}
    \label{fig:framework}
\end{figure}

We answer these questions and use the resulting insights to build a novel programmatic weak supervision system called \textbf{ScriptoriumWS}. A high-level view of ScriptoriumWS is illustrated in Figure \ref{fig:framework}. The system creates LFs by prompting code-generation models to synthesize programs and incorporates them into PWS pipelines. To validate ScriptoriumWS, we conduct experiments with OpenAI Codex \cite{chen2021evaluating}, a state-of-the-art natural language-to-code system based on GPT-3 \cite{brown2020language}. We further propose a complementary approach to incorporate the strength of synthesized and human-designed LFs to improve the performance of the end model.

With the aid of ScriptoriumWS, we explore the advantages that synthesized LFs can bring to the weak supervision framework. 
We study various prompting strategies to gain insight into how to best generate high-quality LFs. 
We conduct experiments in diverse text domains and empirically demonstrate the effectiveness of ScriptoriumWS. 
Excitingly, we find that compared to the human-designed LFs in WRENCH, LFs generated using ScriptoriumWS achieve much higher coverage (the fraction of data points that receive labels) while maintaining high accuracy. 
For example, using the WRENCH benchmark \cite{zhang2021wrench} for comparison, we improve the coverage for the SMS dataset from 40.5\% to 100\% and for the Spouse dataset from 25.8\% to 100\%, while also improving downstream performance by 4.0\% and 5.0\% F1 points, respectively. 

\section{Related Work}

\paragraph{Programmatic Weak Supervision (PWS):} 
PWS refers to a broad set of techniques where the data is labeled using cheaply available but potentially noisy labeling information. This information could be from external knowledge bases, heuristics, web search results, and more. Programmatic weak supervision \cite{ratner2016data, ratner2017snorkel} abstracts out these sources as user-provided (written) labeling functions and gives principled ways to aggregate their outputs to produce accurate pseudolabels.
This framework is practically effective and widely used in industry \cite{ratner2017snorkel,bach2019snorkel} and also offers theoretical guarantees, including consistent estimation of accuracies of labeling functions \cite{ratner2016data, vishwakarma2022Lifting}.
Its main downside is that writing, iterating, and debugging programmatic labeling functions is slow and expensive. 

\paragraph{Automated Weak Supervision (AutoWS):} AutoWS is a class of  techniques that reduce the need for humans to design LFs. 
In many cases, designing LFs can be expensive or challenging, particularly when the feature space is too complex, nuanced, or high-dimensional to be reasoned-about by a human, such as in image and video domains. AutoWS techniques can be used in these situations to automate the LF design process by instead using small models as LFs \cite{varma2018snuba, das2020goggles, boecking2021interactive}, or by augmenting a few given human-designed LFs to explore more rules \cite{inproceedings}. Similarly, it is possible to directly query large pretrained models for noisy label estimates~\cite{smith2022language}. The downside of these approaches is that the resulting labeling functions are typically no longer programs that can be debugged, modified, and re-used. 

\paragraph{Large Pretrained Models and Prompt Engineering:} 
%\bh{Harit's Help}
%Large pretrained models are huge models with billions of parameters trained on enormous corpora. 
%
%Such large-scale training along with features of model architecture gives these models capabilities to answer complex questions, write code, etc. 
Prompting is a common way to tap into the knowledge and capabilities  of large pre-trained models \cite{liu2023PromptingSurvey}. Prompting refers to giving natural language instructions to the model in order to get the answer. These prompts can also contain examples of input-output pairs -- usually referred to as in-context learning \cite{brown2020language, ICL2023Survey}. Prompting has been successfully applied in various applications and understanding various aspects of prompting is a very active area of research. 
% discuss some prompting methods
There are various methods proposed for creating good prompts e.g. 
\cite{arora2023ask} give a general prompting method, chain of thought  prompting \cite{wei2022chain-of-thought} and methods to automate prompt generation \cite{zhou2022-auto-prompt}. For code generation, prompts with detailed instructions, problem statements, partial code, etc. have been used \cite{sami2022-prompting-code, paul2022prompting-code}. We are inspired by these strategies when designing our proposed system. 

%soft-prompting
%\cite{qin2021-soft-prompt},

%prompting debate \cite{mishra2022reframing,wu2022Prompt} 
%\cite{scao2021-prompt-worth}
%\cite{webson-pavlick-2022-prompt}

\textbf{Using Large Pretrained Models for Data Annotation:} Using large language models (LLMs) or other large pretrained models with appropriate prompts to annotate data is a promising direction that can reduce the cost and human effort in data labeling \cite{smith2022language, wang-etal-2021-want-reduce}. %Some of the recent works \cite{ smith2022language,wang-etal-2021-want-reduce} have used LLMs + prompting to annotate data. %According to \cite{smith2022language}, GPT-3 can be incorporated into PWS by modeling diverse prompts and taking returned answers as noisy weak labels. For instance, a data example from the spam dataset can be given to GPT-3 with a prompt like "Is the following comment spam? [Text]", then the prediction can be used as the weak label. While the use of natural language prompts makes PWS easier to take advantage of the knowledge learned by large-scale training in LLMs, this approach has 
The main limitation here is in terms of scalability and privacy. Inference via querying an API for every data example becomes cost-prohibitive when dealing with large-size training datasets, and sending training data through APIs to other organizations poses a risk of privacy leaks, especially for sensitive data.

% \cite{chen2022shoring} pre-trained models + no prompting. 

%% Another approach to generate weak labels is to leverage existing  large language models (LLMs) (i.e., BERT \cite{devlin2018bert}, GPT-3 \cite{brown2020language}, and T0++ \cite{sanh2022multitask}) as labeling sources \cite{chen2022shoring, smith2022language}. According to \cite{smith2022language}, GPT-3 can be incorporated into PWS by modeling diverse prompts and taking returned answers as noisy weak labels. For instance, a data example from the spam dataset can be given to GPT-3 with a prompt like "Is the following comment spam? [Text]", then the prediction can be used as the weak label. While the use of natural language prompts makes PWS easier to take advantage of the knowledge learned by large-scale training in LLMs, this approach has limitations in terms of scalability and privacy. Querying the API for every data example becomes cost-prohibitive when dealing with large-size training datasets, and sending training data through APIs to other organizations poses a risk of privacy leaks, especially for sensitive data.
%% soft-prompt
%% in-context learning

%Besides AutoWS, PromptWS is another approach that makes existing large language models (LLMs) (i.e. GPT-3 \cite{brown2020language} and T0++ \cite{sanh2022multitask}) as labeling sources and queries training data via multiple diverse prompts and model returned answers to generate labels \cite{smith2022language}. 
%
%\paragraph{Code Generation Models}
%% Program Synthesis
%% Codegen
%% Codex (plus competitors, Copilot, etc.) 

\section{Methodology}
In this section, we first describe the programmatic weak supervision (PWS) setup and then discuss approaches that we generate labeling functions by proposing different types of prompts to direct LLMs like Codex in ScriptoriumWS.

\subsection{Programmatic Weak Supervision Setup}
Let $\mathcal{X}, \mathcal{Y}$ be the instance and label spaces,  respectively. 
For each of the $n$ unlabeled examples, $x_i \in \mathcal{X}$,  we observe noisy labels $\lambda_{1,i}, \ldots, \lambda_{m,i}$. 
These are the outputs of $m$ \emph{labeling functions} (LFs) $s_a$, where $s_a : \mathcal{X} \rightarrow \mathcal{Y}$ and $\lambda_{a,i} = s_a(x_i)$. 
These LF outputs are fed to  a two-step process to construct pseudo labels. 
Firstly, we learn a \emph{noise model} (also called a label model) that determines how accurate the sources are. That is, we must learn $\mathbf{\theta}$ for $P_{\mathbf{\theta}}(\lambda_{1}, \lambda_{2}, \ldots, \lambda_m, y)$. 
Note that the model involves true labels $y$ that are not observed for any of the samples and this makes the estimation process challenging. 
Then, pseudo labels for each $x_i$ are inferred using the learned noise model. In other words, we compute $\tilde{y} = \argmax_{y\in\mathcal{Y}} P_{\hat{\mathbf{\theta}}}(\tilde{y}|\lambda_{1}, \lambda_{2}, \ldots, \lambda_m)$. 
Finally, an end model can be trained using the generated training dataset: $D=\{(x_i, \tilde{y}_i)\}_{i=1}^n \subseteq \mathcal{X} \times \mathcal{Y}$. 

A variety of label models are used for the estimation and inference sets.
In this work, we focus on LF generation and use standard label models such as \cite{ratner2019training} and \cite{fu2020fast}.

% \begin{equation}
% P_{\mathbf{\theta}}(\lambda_1, \ldots, \lambda_m|Y=y) = \frac{1}{Z} \exp\left(-\sum_{a=1}^m \theta_a d^2_\mathcal{Y}(\lambda_a,y) { - \sum_{(a,b) \in E} \theta_{a,b} d^2_{\mathcal{Y}}(\lambda_a, \lambda_b)}\right). 
% \label{eq:lbl-model-orig-a}
% \end{equation}

% Here $d_{\mathcal{Y}}$ is a distance function on the label space $\mathcal{Y}$, $Z$ is the normalizing partition function, $\mathbf{\theta} = [\theta_1, \ldots, \theta_m]^T > 0$ are the \emph{canonical} parameters, and $E$ is a set of correlations. Intuitively, if $\theta_a$ is large, the typical distance from $d_a$ to $y$ is small and the LF is reliable; if $\theta_a$ is small, the LF is unreliable. There are several reasons why this model is suitable. Firstly, it belongs to the exponential family and has desirable theoretical characteristics. Additionally, it encompasses well-known forms of noise, such as zero-mean multivariate Gaussian noise for regression and a similar version of the Ising model for binary label cases. 

%Several techniques have been proposed to estimate $\hat{\mathbf{\theta}}$ in order to construct pseudo labels. One way to get such pseudo labels is to compute
%$\tilde{y} = \argmin_{z \in \mathcal{Y}} 1/m \sum_{a=1}^m \hat{\theta}_a d^2_{\mathcal{Y}}(z, \lambda_{a})$. Note the estimated parameters $\hat{\theta}_a$ are used to weight the labeling functions appropriately to  ensure that more reliable LFs receive a larger weight. 

\subsection{ScriptoriumWS System}
ScriptoriumWS is built on top of the PWS framework. Instead of writing LFs $\lambda$ manually, we synthesize them using OpenAI Codex. Codex is a descendant of the GPT-3 model, fine-tuned for use in programming applications. It has shown remarkable performance \cite {xu2022systematic} on code generation tasks across various programming languages. We use the Codex API with natural language prompts to generate code. We vary the temperature parameter from 0 to 0.2 to increase the diversity of the outputs. We feed the synthesized LFs into the PWS pipeline.

\subsection{Types of Prompt}
\begin{figure}
    \centering
    \includegraphics[width=\linewidth]{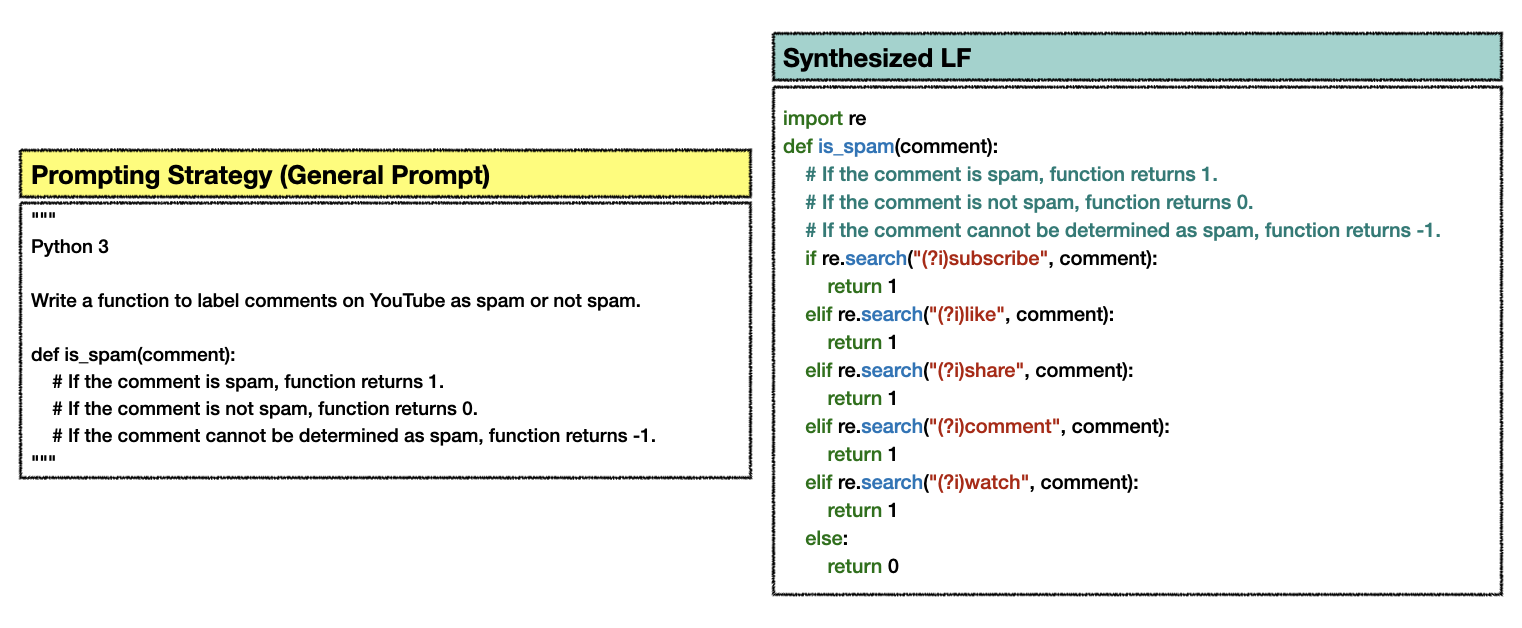}
    \caption{An example of synthesized LF using general prompt strategy for the YouTube spam classification task.}
    \label{fig:gq}
\end{figure}

We explored a variety of prompting strategies, based on the kinds of information typically available to weak supervision users. 
We describe these strategies as being one of five categories, generally going from the least to the most expensive information requirements.

\paragraph{General Prompts:}
%% programming language, task description, function signature, function instructions
%To make users easy to write prompt for diverse datasets, prompt should have specification and adaptability. In ScriptorimWS, we 
First, we propose a general prompt format that can be easily extended with additional information. A general prompt includes four components, which are the use of programming language, basic task description, function signature, and labeling instructions. We demonstrate an example for the YouTube spam classification task \cite{alberto2015tubespam} in Figure \ref{fig:gq}. 

A general prompt first provides the programming language to be used to synthesize code. 
%
%It helps to ensure the consistency of returned code. 
Next, the basic task description provides an overview of what the function is expected to do.
Afterward, the function signature outlines the name of the synthesized program and the input that the code generation model should use.
Finally, we place labeling instructions into the function signature to specify the format and structure of the returned output.

%% \begin{figure}
%%     \centering
%%     \includegraphics[width=\linewidth]{figures/mission_statement.png}
%%     \caption{An example of synthesized LF using mission statement strategy for the YouTube spam classification task}
%%     \label{fig:ms}
%% \end{figure}

\paragraph{Mission Statement:}
In addition to providing a basic task description, we also propose an extended type of prompt, which we call the \emph{Mission Statement}. Here we add information to the general prompt to give the code generation model a better understanding of the task in the context. This additional information includes the use of external knowledge bases, and it can include relevant background descriptions about the problem or high-level dataset information (i.e., feature or label classes).

\begin{figure}
    \centering
    \includegraphics[width=\linewidth]{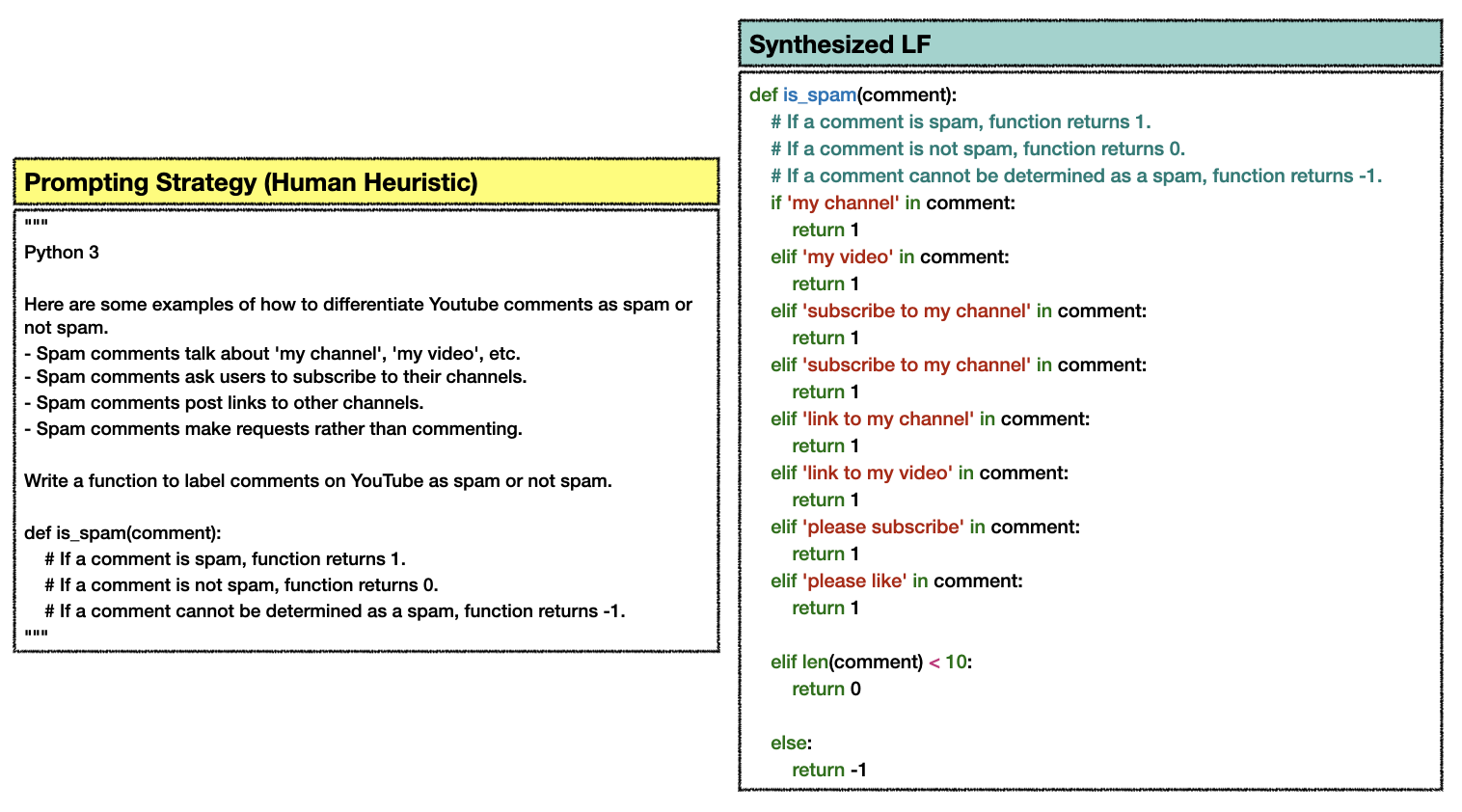}
    \caption{An example of a synthesized LF by using human heuristic strategy for the YouTube spam classification task}
    \label{fig:hh}
\end{figure}

\paragraph{Human Heuristic:}
In practical applications, users generally have a wealth of prior knowledge and expertise that they can bring to the prompt, including heuristic rules and domain-specific knowledge. Incorporating this prior knowledge into the prompt can be helpful in guiding the code generation model to have a better understanding of the problem and potentially develop a more effective solution that leverages the user's expertise. For example, if a user knows that certain keywords are indicative of spam, they could include this information in the prompt. In ScriptoriumWS, we reference keywords from existing human-designed LFs and write them into heuristic rules then add these rules to the prompt. We demonstrate an example in the category of human heuristic in Figure \ref{fig:hh}.

\begin{figure}[t!]
    \centering
    \includegraphics[width=\linewidth]{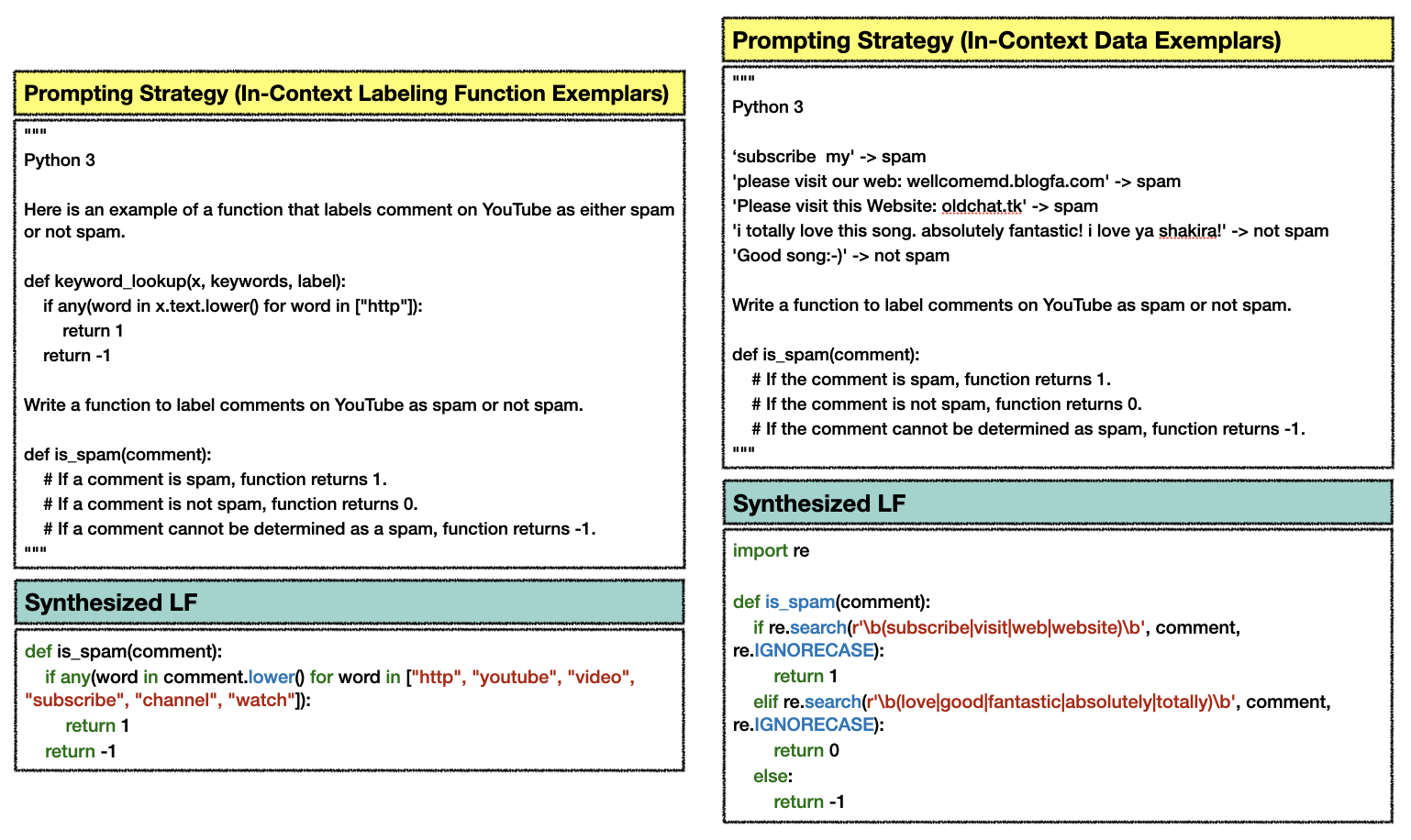}
    \caption{Two synthesized LF examples generated by adding label function examples (left) and data examples (right) for the YouTube spam classification task. We can see that code generation model takes the given label function example as reference and learn the relationship between data examples and their expected outputs to extend and synthesize it own program.}
    \label{fig:lfe}
\end{figure}

\paragraph{In-Context Labeling Function Exemplars:}
In-context few-shot learning is a popular approach to perform a new task by inputting a few examples without the need of fine-tuning. We consider a practical scenario where users have already written some LFs or are allowed to access a few existing LFs. Such LFs can be incorporated into the prompt. The code generation model can use them as function templates to synthesize its own LF, which can be more closely aligned with the user's prior knowledge and expertise, rather than relying solely on the model's own training data.

%% \begin{figure}
%%     \centering
%%     \includegraphics[width=\linewidth]{figures/data_example.png}
%%     \caption{A prompt example using type of data example for YouTube dataset.}
%%     \label{fig:de}
%% \end{figure}

\paragraph{In-Context Data Exemplars:}
Besides providing Codex with heuristic rules and in-context few-shot learning with human-designed LFs, we propose another approach by incorporating a few labeled data examples into the prompt to direct the model to understand the problem. Given data examples can serve as concrete illustrations of the problem and provide a clearer understanding of the task and the expected output. This can be especially easy and useful when the problem domain is too complex to design heuristic rules or labeling functions manually.

\section{Experiments}
In this section, we validate the capability of the proposed system. 
We implement ScriptoriumWS on the top of weak supervision pipeline proposed as part of the WRENCH benchmark \cite{zhang2021wrench} and use synthesized LFs to generate weak labels to learn the label model and then subsequently the end model.

\subsection{Setup}

\paragraph{Datasets}
We evaluate our approach using four different types of text tasks involving a set of 6 datasets originally included in WRENCH. These 6 datasets are the IMDb \cite{ren2020denoising} and Yelp \cite{ren2020denoising} datasets for sentiment classification, the YouTube \cite{alberto2015tubespam} and SMS \cite{almeida2011contributions} datasets for spam classification, the AGNews \cite{ren2020denoising} dataset for topic classification, and the Spouse \cite{ratner2017snorkel} dataset for relation classification. 

\paragraph{Label Model \& End Model}
%% Label models: Snorkel, WMV, MV, DS, and FS.
%% End model: logistic regression.
Our system is compatible with any choice of label and end model.
For ease of comparison, we follow WRENCH and evaluate with five label models to aggregate the output of our synthesized LFs: majority vote (MV), weighted majority vote (WMV), Snorkel \citep{ratner2017snorkel}, Dawid-Skene (DS) \cite{Dawid:Skene:79}, and FlyingSquid (FS) \cite{threerius}. 
Finally, once we generate labeled training datasets using these label models alongside our LFs, we train a downstream model---for the sake of simplicity, we use logistic regression as end model for all tasks. 

\subsection{Analysis}
We use our evaluation platform to validate the following claims:

\begin{table}[t!]
\centering
\resizebox{\textwidth}{!}{%
\begin{tabular}{@{}l|ccccc|ccccc@{}}
\toprule
 & \multicolumn{5}{c|}{\textbf{IMDb}} & \multicolumn{5}{c}{\textbf{Yelp}} \\ \cmidrule(l){2-11} 
 & \#LFs & \begin{tabular}[c]{@{}c@{}}Avg. \\ Coverage\end{tabular} & \begin{tabular}[c]{@{}c@{}}Avg. \\ Overlap\end{tabular} & \begin{tabular}[c]{@{}c@{}}Avg. \\ Conflict\end{tabular} & \begin{tabular}[c]{@{}c@{}}Avg. \\ Accuracy\end{tabular} & \#LFs & \begin{tabular}[c]{@{}c@{}}Avg. \\ Coverage\end{tabular} & \begin{tabular}[c]{@{}c@{}}Avg. \\ Overlap\end{tabular} & \begin{tabular}[c]{@{}c@{}}Avg. \\ Conflict\end{tabular} & \begin{tabular}[c]{@{}c@{}}Avg. \\ Accuracy\end{tabular} \\ \midrule
WRENCH & 5 & 0.236 & 0.116 & 0.045 & 0.699 & 8 & 0.183 & 0.136 & 0.049 & 0.731 \\ \midrule
General Prompt & 6 & 0.894 & 0.887 & 0.331 & 0.595 & 11 & 0.716 & 0.716 & 0.213 & 0.736 \\
+ Mission Statement & 5 & 0.780 & 0.766 & 0.609 & 0.568 & 7 & 0.697 & 0.689 & 0.168 & 0.686 \\
+ Human Heuristic & 6 & 0.764 & 0.758 & 0.596 & 0.644 & 5 & 0.783 & 0.774 & 0.088 & 0.658 \\
+ Labeling Function Exemplars & 5 & 0.805 & 0.792 & 0.133 & 0.593 & 5 & 0.814 & 0.812 & 0.258 & 0.690 \\
+ Data Exemplars & 5 & 0.895 & 0.895 & 0.382 & 0.633 & 6 & 0.701 & 0.689 & 0.109 & 0.702 \\ \midrule
 & \multicolumn{5}{c|}{\textbf{SMS}} & \multicolumn{5}{c}{\textbf{YouTube}} \\ \cmidrule(l){2-11} 
 & \#LFs & \begin{tabular}[c]{@{}c@{}}Avg. \\ Coverage\end{tabular} & \begin{tabular}[c]{@{}c@{}}Avg. \\ Overlap\end{tabular} & \begin{tabular}[c]{@{}c@{}}Avg. \\ Conflict\end{tabular} & \begin{tabular}[c]{@{}c@{}}Avg. \\ Accuracy\end{tabular} & \#LFs & \begin{tabular}[c]{@{}c@{}}Avg. \\ Coverage\end{tabular} & \begin{tabular}[c]{@{}c@{}}Avg. \\ Overlap\end{tabular} & \begin{tabular}[c]{@{}c@{}}Avg. \\ Conflict\end{tabular} & \begin{tabular}[c]{@{}c@{}}Avg. \\ Accuracy\end{tabular} \\ \midrule
WRENCH & 73 & 0.007 & 0.003 & 0.000 & 0.973 & 10 & 0.170 & 0.132 & 0.075 & 0.826 \\ \midrule
General Prompt & 8 & 0.815 & 0.815 & 0.260 & 0.897 & 9 & 0.592 & 0.592 & 0.493 & 0.646 \\
+ Mission Statement & 8 & 0.819 & 0.819 & 0.324 & 0.817 & 9 & 0.643 & 0.643 & 0.602 & 0.607 \\
+ Human Heuristic & 9 & 0.741 & 0.741 & 0.118 & 0.821 & 8 & 0.570 & 0.570 & 0.491 & 0.802 \\
+ Labeling Function Exemplars & 8 & 0.038 & 0.014 & 0.001 & 0.822 & 6 & 0.662 & 0.662 & 0.349 & 0.795 \\
+ Data Exemplars & 8 & 0.612 & 0.612 & 0.366 & 0.749 & 8 & 0.534 & 0.534 & 0.397 & 0.793 \\ \midrule
 & \multicolumn{5}{c|}{\textbf{Spouse}} & \multicolumn{5}{c}{\textbf{AGNews}} \\ \cmidrule(l){2-11} 
 & \#LFs & \begin{tabular}[c]{@{}c@{}}Avg. \\ Coverage\end{tabular} & \begin{tabular}[c]{@{}c@{}}Avg. \\ Overlap\end{tabular} & \begin{tabular}[c]{@{}c@{}}Avg. \\ Conflict\end{tabular} & \begin{tabular}[c]{@{}c@{}}Avg. \\ Accuracy\end{tabular} & \#LFs & \begin{tabular}[c]{@{}c@{}}Avg. \\ Coverage\end{tabular} & \begin{tabular}[c]{@{}c@{}}Avg. \\ Overlap\end{tabular} & \begin{tabular}[c]{@{}c@{}}Avg. \\ Conflict\end{tabular} & \begin{tabular}[c]{@{}c@{}}Avg. \\ Accuracy\end{tabular} \\ \midrule
WRENCH & 9 & 0.042 & 0.021 & 0.009 & 0.586 & 9 & 0.103 & 0.051 & 0.024 & 0.817 \\ \midrule
General Prompt & 8 & 1.000 & 1.000 & 0.324 & 0.807 & 8 & 0.305 & 0.279 & 0.080 & 0.565 \\
+ Mission Statement & 9 & 0.279 & 0.208 & 0.168 & 0.404 & 4 & 0.373 & 0.215 & 0.123 & 0.338 \\
+ Human Heuristic & 8 & 0.295 & 0.264 & 0.050 & 0.456 & 4 & 0.346 & 0.327 & 0.064 & 0.818 \\
+ Labeling Function Exemplars & 5 & 0.417 & 0.307 & 0.023 & 0.444 & 8 & 0.481 & 0.472 & 0.191 & 0.530 \\
+ Data Exemplars & 8 & 0.601 & 0.601 & 0.240 & 0.595 & 5 & 0.345 & 0.244 & 0.107 & 0.636 \\ \bottomrule
\end{tabular}}
\centering\caption{Statistics of synthesized LFs.}
\label{basic stats}
\end{table}

%\paragraph{How effective are our synthesized LFs?}
\paragraph{Are ScriptoriumWS LFs comparable to human-designed LFs?}
%% performance is comparable to humanLF
%% show LF performance: accuracy, coverage, overlap, conflict (table)
%To validate the effectiveness of ScriptoriumWS, 
We hypothesize that the synthesized LFs generated by ScriptoriumWS can provide results that are comparable to human-designed LFs. 
To see the strengths and weaknesses of synthesized LFs, we include four basic measurements for LFs generated by different prompting strategies.
These are coverage, overlap, conflict, and accuracy. 
Coverage is the fraction of the dataset labeled by a given LF. 
Overlap shows the fraction of the dataset with at least two (non-abstain) labels. 
Conflict indicates a data example for which at least one other LF provides a different estimate. 
Accuracy computes the fraction of the correctly labeled dataset. 
We take the average of these indicators over our synthesized LFs and compare them with human-designed LFs in WRENCH. 

The results are shown in Table \ref{basic stats}. 
They demonstrate that ScriptoriumWS is capable of generating LFs that have comparable accuracy to human-designed LFs. 
We observe that synthesized LFs significantly outperform human-designed LFs in terms of coverage. This is not surprising, as human-crafted LFs are often very specific and cannot cover too much of the dataset.  
On the other hand, we see that there exist more conflicts among outputs produced by synthesized LFs. 
However, this is not a concern, as such conflicts are resolved by (and in fact, are useful to learn) the label model.

\begin{table}[t!]
\resizebox{\linewidth}{!}{
\begin{tabular}{@{}l|cccccccccccccc@{}}
\toprule
 & \multicolumn{7}{c|}{\textbf{IMDb (Accuracy)}} & \multicolumn{7}{c}{\textbf{Yelp (Accuracy)}} \\ \cmidrule(l){2-15} 
 & Snorkel & WMV & MV & DS & FS & Avgerage & \multicolumn{1}{c|}{Coverage} & Snorkel & WMV & MV & DS & FS & Avgerage & Coverage \\ \midrule
WRENCH & 0.701 & 0.710 & 0.710 & 0.706 & 0.704 & 0.706 & \multicolumn{1}{c|}{0.876} & 0.690 & 0.685 & 0.702 & 0.715 & 0.687 & 0.696 & 0.828 \\ \midrule
General Prompt & 0.661 & 0.587 & 0.587 & 0.559 & 0.606 & 0.600 & \multicolumn{1}{c|}{\textbf{0.998}} & 0.766 & 0.693 & 0.703 & 0.748 & 0.700 & \textbf{0.722} & \textbf{0.991} \\
+ Mission Statement & 0.613 & 0.600 & 0.600 & 0.542 & 0.612 & 0.593 & \multicolumn{1}{c|}{\textbf{0.973}} & 0.743 & 0.665 & 0.675 & 0.634 & 0.670 & 0.677 & \textbf{0.988} \\
+ Human Heuristic & 0.710 & 0.652 & 0.652 & 0.588 & 0.649 & 0.650 & \multicolumn{1}{c|}{\textbf{0.985}} & 0.661 & 0.635 & 0.642 & 0.695 & 0.610 & 0.648 & \textbf{0.955} \\
+ Labeling Function Exemplars & 0.650 & 0.614 & 0.614 & 0.596 & 0.612 & 0.617 & \multicolumn{1}{c|}{\textbf{0.941}} & 0.793 & 0.670 & 0.692 & 0.728 & 0.729 & \textbf{0.722} & \textbf{0.994} \\
+ Data Exemplars & 0.713 & 0.676 & 0.676 & 0.690 & 0.698 & \textbf{0.691} & \multicolumn{1}{c|}{\textbf{1.000}} & 0.766 & 0.678 & 0.688 & 0.736 & 0.685 & \textbf{0.711} & \textbf{0.990} \\ \midrule
 & \multicolumn{7}{c}{\textbf{SMS (F1-score)}} & \multicolumn{7}{c}{\textbf{YouTube (Accuracy)}} \\ \cmidrule(l){2-15} 
 & Snorkel & WMV & MV & DS & FS & Avgerage & Coverage & Snorkel & WMV & MV & DS & FS & Avgerage & Coverage \\ \midrule
WRENCH & 0.048 & 0.240 & 0.240 & 0.049 & 0.000 & 0.115 & \multicolumn{1}{c|}{0.405} & 0.852 & 0.780 & 0.840 & 0.832 & 0.764 & 0.814 & 0.893 \\ \midrule
General Prompt & 0.632 & 0.526 & 0.672 & 0.622 & 0.632 & \textbf{0.617} & \multicolumn{1}{c|}{\textbf{1.000}} & 0.760 & 0.700 & 0.724 & 0.668 & 0.784 & 0.727 & \textbf{1.000} \\
+ Mission Statement & 0.599 & 0.029 & 0.615 & 0.599 & 0.599 & \textbf{0.488} & \multicolumn{1}{c|}{\textbf{1.000}} & 0.540 & 0.624 & 0.648 & 0.688 & 0.468 & 0.594 & \textbf{1.000} \\
+ Human Heuristic & 0.606 & 0.412 & 0.554 & 0.529 & 0.536 & \textbf{0.527} & \multicolumn{1}{c|}{\textbf{1.000}} & 0.556 & 0.740 & 0.748 & 0.748 & 0.776 & 0.714 & \textbf{1.000} \\
+ Labeling Function Exemplars & 0.086 & 0.317 & 0.317 & 0.237 & 0.027 & \textbf{0.197} & \multicolumn{1}{c|}{0.218} & 0.740 & 0.740 & 0.740 & 0.748 & 0.740 & 0.742 & \textbf{1.000} \\
+ Data Exemplars & 0.650 & 0.337 & 0.628 & 0.640 & 0.630 & \textbf{0.577} & \multicolumn{1}{c|}{\textbf{1.000}} & 0.888 & 0.844 & 0.868 & 0.728 & 0.888 & \textbf{0.843} & \textbf{1.000} \\ \midrule
 & \multicolumn{7}{c|}{\textbf{Spouse (F1-score)}} & \multicolumn{7}{c}{\textbf{AGNews (Accuracy)}} \\ \cmidrule(l){2-15} 
 & Snorkel & WMV & MV & DS & FS & Avgerage & \multicolumn{1}{c|}{Coverage} & Snorkel & WMV & MV & DS & FS & Avgerage & Coverage \\ \midrule
WRENCH & 0.498 & 0.205 & 0.208 & 0.155 & 0.343 & 0.282 & \multicolumn{1}{c|}{0.258} & 0.625 & 0.640 & 0.638 & 0.628 & 0.610 & 0.628 & 0.691 \\ \midrule
General Prompt & 0.395 & 0.090 & 0.387 & 0.382 & 0.374 & \textbf{0.325} & \multicolumn{1}{c|}{\textbf{1.000}} & 0.537 & 0.530 & 0.529 & 0.410 & 0.544 & 0.510 & \textbf{0.692} \\
+ Mission Statement & 0.381 & 0.173 & 0.355 & 0.399 & 0.345 & \textbf{0.331} & \multicolumn{1}{c|}{\textbf{1.000}} & 0.397 & 0.393 & 0.347 & 0.372 & 0.372 & 0.376 & \textbf{1.000} \\
+ Human Heuristic & 0.393 & 0.204 & 0.243 & 0.391 & 0.340 & \textbf{0.315} & \multicolumn{1}{c|}{\textbf{0.470}} & 0.597 & 0.580 & 0.572 & 0.536 & 0.597 & \textbf{0.576} & 0.667 \\
+ Labeling Function Exemplars & 0.394 & 0.172 & 0.169 & 0.165 & 0.287 & 0.237 & \multicolumn{1}{c|}{\textbf{1.000}} & 0.544 & 0.527 & 0.525 & 0.485 & 0.525 & 0.521 & \textbf{0.811} \\
+ Data Exemplars & 0.395 & 0.134 & 0.378 & 0.389 & 0.383 & \textbf{0.336} & \multicolumn{1}{c|}{\textbf{1.000}} & 0.477 & 0.458 & 0.421 & 0.404 & 0.471 & 0.446 & \textbf{1.000} \\ \bottomrule
\end{tabular}}
\centering\caption{Performance of label models across different type of prompting strategies.}
\label{label model perf}
\end{table}

%\paragraph{How does synthesized LFs perform on the label model?}
\paragraph{How does ScriptoriumWS perform in PWS pipelines?}
%Previously, we validate that ScriptoriumWS has the potential to be a useful tool for reducing the human effort required in designing LFs. 
%Next, we expect to see the capability of synthesized LFs in the PWS pipelines. 
We anticipate that as LFs from ScriptoriumWS are comparable to human-designed LFs, such LFs will yield good performance in downstream tasks.  
We train the label model to aggregate the outputs of synthesized LFs and evaluate the performance of the label model on the testing dataset.
We compute model performance across different label models for each type of prompting strategy.

The results are shown in Table \ref{label model perf}. We find that the performance of the label model using synthesized LFs is generally on par with that of the label model using human-designed LFs while achieving much higher coverage. 
In particular, coverage on the SMS and Spouse datasets are low when using human-designed LFs from WRENCH; however, when using our synthesized LFs, \textit{we achieve $100\%$ coverage while also achieving higher F1-scores.} 
These results suggest that synthesized LFs can be a valuable resource for PWS pipelines and provide strong evidence for the efficacy of ScriptoriumWS in practical applications.

\begin{table}[h!]
\resizebox{\linewidth}{!}{
\begin{tabular}{@{}l|cccccccc|cccccccc@{}}
\toprule
 & \multicolumn{8}{c|}{\textbf{IMDb (Accuracy)}} & \multicolumn{8}{c}{\textbf{Yelp (Accuracy)}} \\ \cmidrule(l){2-17} 
 & \#LFs & \begin{tabular}[c]{@{}c@{}}Snorkel \\ + LR\end{tabular} & \begin{tabular}[c]{@{}c@{}}WMV \\ + LR\end{tabular} & \begin{tabular}[c]{@{}c@{}}MV \\ + LR\end{tabular} & \begin{tabular}[c]{@{}c@{}}DS \\ + LR\end{tabular} & \begin{tabular}[c]{@{}c@{}}FS \\ + LR\end{tabular} & Average & Coverage & \#LFs & \begin{tabular}[c]{@{}c@{}}Snorkel \\ + LR\end{tabular} & \begin{tabular}[c]{@{}c@{}}WMV \\ + LR\end{tabular} & \begin{tabular}[c]{@{}c@{}}MV \\ + LR\end{tabular} & \begin{tabular}[c]{@{}c@{}}DS \\ + LR\end{tabular} & \begin{tabular}[c]{@{}c@{}}FS \\ + LR\end{tabular} & Average & Coverage \\ \midrule
WRENCH & 5 & 0.758 & 0.754 & 0.754 & 0.754 & 0.756 & 0.755 & 0.876 & 8 & 0.722 & 0.649 & 0.694 & 0.807 & 0.737 & 0.722 & 0.828 \\
+ General Prompt & +6 & 0.740 & 0.737 & 0.742 & 0.767 & 0.739 & 0.745 & \textbf{1.000} & +11 & 0.750 & 0.671 & 0.704 & 0.801 & 0.730 & \textbf{0.731} & \textbf{1.000} \\
+ Mission Statement & +5 & 0.732 & 0.756 & 0.761 & 0.767 & 0.747 & 0.753 & \textbf{1.000} & +7 & 0.734 & 0.660 & 0.693 & 0.815 & 0.753 & \textbf{0.731} & \textbf{1.000} \\
+ Human Heuristic & +6 & 0.763 & 0.769 & 0.771 & 0.785 & 0.757 & \textbf{0.769} & \textbf{1.000} & +5 & 0.680 & 0.619 & 0.661 & 0.804 & 0.703 & 0.693 & \textbf{1.000} \\
+ Labeling Function Exemplars & +5 & 0.737 & 0.746 & 0.750 & 0.786 & 0.735 & 0.751 & \textbf{1.000} & +5 & 0.656 & 0.664 & 0.706 & 0.808 & 0.711 & 0.709 & \textbf{1.000} \\
+ Data Exemplars & +5 & 0.770 & 0.752 & 0.757 & 0.758 & 0.767 & \textbf{0.761} & \textbf{1.000} & +6 & 0.724 & 0.656 & 0.705 & 0.821 & 0.748 & \textbf{0.731} & \textbf{1.000} \\ \midrule
 & \multicolumn{8}{c|}{\textbf{SMS (F1-score)}} & \multicolumn{8}{c}{\textbf{YouTube (Accuracy)}} \\ \cmidrule(l){2-17} 
 & \#LFs & \begin{tabular}[c]{@{}c@{}}Snorkel \\ + LR\end{tabular} & \begin{tabular}[c]{@{}c@{}}WMV \\ + LR\end{tabular} & \begin{tabular}[c]{@{}c@{}}MV \\ + LR\end{tabular} & \begin{tabular}[c]{@{}c@{}}DS \\ + LR\end{tabular} & \begin{tabular}[c]{@{}c@{}}FS \\ + LR\end{tabular} & Average & Coverage & \#LFs & \begin{tabular}[c]{@{}c@{}}Snorkel \\ + LR\end{tabular} & \begin{tabular}[c]{@{}c@{}}WMV \\ + LR\end{tabular} & \begin{tabular}[c]{@{}c@{}}MV \\ + LR\end{tabular} & \begin{tabular}[c]{@{}c@{}}DS \\ + LR\end{tabular} & \begin{tabular}[c]{@{}c@{}}FS \\ + LR\end{tabular} & Average & Coverage \\ \midrule
WRENCH & 73 & 0.678 & 0.772 & 0.756 & 0.750 & 0.057 & 0.603 & 0.405 & 10 & 0.808 & 0.732 & 0.808 & 0.828 & 0.788 & 0.793 & 0.893 \\
+ General Prompt & +8 & 0.720 & 0.542 & 0.750 & 0.709 & 0.473 & \textbf{0.639} & \textbf{1.000} & +9 & 0.832 & 0.740 & 0.788 & 0.820 & 0.756 & 0.787 & \textbf{1.000} \\
+ Mission Statement & +8 & 0.619 & 0.405 & 0.672 & 0.576 & 0.420 & 0.538 & \textbf{1.000} & +9 & 0.820 & 0.732 & 0.756 & 0.784 & 0.716 & 0.762 & \textbf{1.000} \\
+ Human Heuristic & +9 & 0.632 & 0.476 & 0.582 & 0.594 & 0.482 & 0.553 & \textbf{1.000} & +8 & 0.808 & 0.756 & 0.808 & 0.816 & 0.772 & \textbf{0.792} & \textbf{1.000} \\
+ Labeling Function Exemplars & +8 & 0.405 & 0.465 & 0.473 & 0.610 & 0.029 & 0.396 & \textbf{1.000} & +6 & 0.776 & 0.756 & 0.780 & 0.796 & 0.780 & 0.778 & \textbf{1.000} \\
+ Data Exemplars & +8 & 0.692 & 0.418 & 0.746 & 0.722 & 0.509 & \textbf{0.617} & \textbf{1.000} & +8 & 0.800 & 0.756 & 0.800 & 0.780 & 0.788 & 0.785 & \textbf{1.000} \\ \midrule
 & \multicolumn{8}{c|}{\textbf{Spouse (F1-score)}} & \multicolumn{8}{c}{\textbf{AGNews (Accuracy)}} \\ \cmidrule(l){2-17} 
 & \#LFs & \begin{tabular}[c]{@{}c@{}}Snorkel \\ + LR\end{tabular} & \begin{tabular}[c]{@{}c@{}}WMV \\ + LR\end{tabular} & \begin{tabular}[c]{@{}c@{}}MV \\ + LR\end{tabular} & \begin{tabular}[c]{@{}c@{}}DS \\ + LR\end{tabular} & \begin{tabular}[c]{@{}c@{}}FS \\ + LR\end{tabular} & Average & Coverage & \#LFs & \begin{tabular}[c]{@{}c@{}}Snorkel \\ + LR\end{tabular} & \begin{tabular}[c]{@{}c@{}}WMV \\ + LR\end{tabular} & \begin{tabular}[c]{@{}c@{}}MV \\ + LR\end{tabular} & \begin{tabular}[c]{@{}c@{}}DS \\ + LR\end{tabular} & \begin{tabular}[c]{@{}c@{}}FS \\ + LR\end{tabular} & Average & Coverage \\ \midrule
WRENCH & 9 & 0.220 & 0.179 & 0.181 & 0.166 & 0.268 & 0.203 & 0.258 & 9 & 0.825 & 0.823 & 0.827 & 0.829 & 0.817 & 0.824 & 0.691 \\
+ General Prompt & +8 & 0.157 & 0.298 & 0.303 & 0.301 & 0.155 & \textbf{0.243} & \textbf{1.000} & +8 & 0.806 & 0.823 & 0.825 & 0.751 & 0.817 & 0.804 & \textbf{1.000} \\
+ Mission Statement & +9 & 0.101 & 0.308 & 0.301 & 0.314 & 0.195 & \textbf{0.244} & \textbf{1.000} & +4 & 0.684 & 0.713 & 0.714 & 0.726 & 0.719 & 0.711 & \textbf{1.000} \\
+ Human Heuristic & +8 & 0.058 & 0.213 & 0.218 & 0.299 & 0.104 & 0.178 & \textbf{1.000} & +4 & 0.813 & 0.811 & 0.812 & 0.766 & 0.806 & 0.802 & \textbf{1.000} \\
+ Labeling Function Exemplars & +5 & 0.147 & 0.152 & 0.154 & 0.148 & 0.093 & 0.139 & \textbf{1.000} & +8 & 0.784 & 0.797 & 0.795 & 0.794 & 0.790 & 0.792 & \textbf{1.000} \\
+ Data Exemplars & +8 & 0.192 & 0.301 & 0.300 & 0.308 & 0.164 & \textbf{0.253} & \textbf{1.000} & +5 & 0.711 & 0.726 & 0.725 & 0.712 & 0.737 & 0.722 & \textbf{1.000} \\ \bottomrule
\end{tabular}}
\centering\caption{Performance of end models across different type of prompting strategies.}
\label{end model perf}
\end{table}

\paragraph{How does prompting strategy affect performance?} 
Different prompting strategies can lead to LFs that are more or less aligned with the user's prior knowledge and expertise, which in turn can affect the quality of LFs and their performance in downstream pipelines. 
For instance, providing labeled data in the prompt can prime Codex with information about the relationships between the input features and the target labels. 
On the other hand, providing heuristic rules in the prompt can lead Codex to focus more on the user's prior knowledge. 
We initially hypothesized that these different prompting strategies would lead to a discernible pattern---some strategies would dominate in certain settings. 
However, in our experimental results shown in Table~\ref{label model perf}, suggest no such pattern, leading to an inconclusive result. 
%, and we find that 
%the `correct' prompting method remains unclear, as different prompting methods perform differently across datasets. 
%there is inconclusive finding to indicate which prompting strategy work the best. 
%Instead, the best prompting strategy may depend on the specific task and dataset at hand. 
It is important to carefully consider the goals and requirements of the task and choose a prompt that is suitable for the task. 
A deeper analysis that elucidates the successes and failure modes of each prompting strategy is required to evaluate our original hypothesis and, perhaps more broadly, to better understand the role of prompting in code generation. 
%Further experiments and evaluations are required to better understand an efficient way to choose the appropriate prompting strategy. 

%\paragraph{How can synthesized LFs to be used to improve the end model?}
\paragraph{Can end model performance be improved by combining ScriptoriumWS with PWS?}
%In the end model training, those data points uncovered by LFs may not be included. 
End models can only be trained on points that receive labels. 
Building on our observation that synthesized LFs offer high coverage, we hypothesize that end model performance can be improved over the standard PWS pipeline by simply including the examples that are labeled by ScriptoriumWS.
%
%Based on this idea, we propose a complementary approach to incorporate synthesized LFs and human-designed LFs. 
This yields a complementary approach that incorporates both our synthesized LFs for points that are not labeled by human-designed LFs, and the labels that were originally produced by human-designed LFs. 
%We make synthesized LFs to label those uncovered data points by the human-designed LFs and train label model on two types of outputs respectively. 
We train the end model on the union of these two sets. 
%Ultimately, we combine pseudo labels from two label models, and the end model is trained on this merged generated dataset.
%
%We anticipate observing improved coverage and performance.
In Table \ref{end model perf}, we show the performance of the end model when using this approach.
As before, the dataset is fully covered by this approach and the end-model performance improves due to the significant increase in labeled examples. 
%As expected, we find that the dataset can be fully covered, and the end model performance can be improved by leveraging the strength of synthesized LFs. 
%This finding justifies the notion that the synthesized LFs can be complementary to human-designed LFs, leading to improved end model performance. 
%These findings can be valuable for users who need to label large datasets but have limited resources for writing sufficient labeling function.
%With the aid of ScriptoriumWS, users can quickly and efficiently synthesize LFs to cover the uncovered data points to improve the performance of the end model.
This shows that ScriptoriumWS can be used complementarity with existing PWS pipelines, for which human-designed LFs have already been created to improve performance.

\section{Conclusion}
In this paper, we aim to reduce the human effort required to design weak supervision labeling functions. We propose a novel system, ScriptoriumWS, to leverage code-generation models to provide programming assistance to synthesize labeling functions (LFs) automatically. We study a variety of prompting strategies, propose a simple pipeline, and obtain promising results when comparing to human-designed labeling functions on the WRENCH weak supervision benchmark. Our results show the effectiveness of ScriptoriumWS and can be used as complement approach with existing PWS pipelines to improve end model performance.

\bibliography{reference}

\providecommand{\latin}[1]{#1}
\makeatletter
\providecommand{\doi}
  {\begingroup\let\do\@makeother\dospecials
  \catcode`\{=1 \catcode`\}=2 \doi@aux}
\providecommand{\doi@aux}[1]{\endgroup\texttt{#1}}
\makeatother
\providecommand*\mcitethebibliography{\thebibliography}
\csname @ifundefined\endcsname{endmcitethebibliography}  {\let\endmcitethebibliography\endthebibliography}{}
\begin{mcitethebibliography}{34}
\providecommand*\natexlab[1]{#1}
\providecommand*\mciteSetBstSublistMode[1]{}
\providecommand*\mciteSetBstMaxWidthForm[2]{}
\providecommand*\mciteBstWouldAddEndPuncttrue
  {\def\EndOfBibitem{\unskip.}}
\providecommand*\mciteBstWouldAddEndPunctfalse
  {\let\EndOfBibitem\relax}
\providecommand*\mciteSetBstMidEndSepPunct[3]{}
\providecommand*\mciteSetBstSublistLabelBeginEnd[3]{}
\providecommand*\EndOfBibitem{}
\mciteSetBstSublistMode{f}
\mciteSetBstMaxWidthForm{subitem}{(\alph{mcitesubitemcount})}
\mciteSetBstSublistLabelBeginEnd
  {\mcitemaxwidthsubitemform\space}
  {\relax}
  {\relax}

\bibitem[Ratner \latin{et~al.}(2016)Ratner, De~Sa, Wu, Selsam, and R{\'e}]{ratner2016data}
Ratner,~A.~J.; De~Sa,~C.~M.; Wu,~S.; Selsam,~D.; R{\'e},~C. Data programming: Creating large training sets, quickly. \emph{Advances in neural information processing systems} \textbf{2016}, \emph{29}\relax
\mciteBstWouldAddEndPuncttrue
\mciteSetBstMidEndSepPunct{\mcitedefaultmidpunct}
{\mcitedefaultendpunct}{\mcitedefaultseppunct}\relax
\EndOfBibitem
\bibitem[Ratner \latin{et~al.}(2017)Ratner, Bach, Ehrenberg, Fries, Wu, and R{\'e}]{ratner2017snorkel}
Ratner,~A.; Bach,~S.~H.; Ehrenberg,~H.; Fries,~J.; Wu,~S.; R{\'e},~C. Snorkel: Rapid training data creation with weak supervision. Proceedings of the VLDB Endowment. International Conference on Very Large Data Bases. 2017; p 269\relax
\mciteBstWouldAddEndPuncttrue
\mciteSetBstMidEndSepPunct{\mcitedefaultmidpunct}
{\mcitedefaultendpunct}{\mcitedefaultseppunct}\relax
\EndOfBibitem
\bibitem[Ratner \latin{et~al.}(2019)Ratner, Hancock, Dunnmon, Sala, Pandey, and R{\'e}]{ratner2019training}
Ratner,~A.; Hancock,~B.; Dunnmon,~J.; Sala,~F.; Pandey,~S.; R{\'e},~C. Training complex models with multi-task weak supervision. Proceedings of the AAAI Conference on Artificial Intelligence. 2019; pp 4763--4771\relax
\mciteBstWouldAddEndPuncttrue
\mciteSetBstMidEndSepPunct{\mcitedefaultmidpunct}
{\mcitedefaultendpunct}{\mcitedefaultseppunct}\relax
\EndOfBibitem
\bibitem[Fu \latin{et~al.}(2020)Fu, Chen, Sala, Hooper, Fatahalian, and R{\'e}]{fu2020fast}
Fu,~D.; Chen,~M.; Sala,~F.; Hooper,~S.; Fatahalian,~K.; R{\'e},~C. Fast and three-rious: Speeding up weak supervision with triplet methods. International Conference on Machine Learning. 2020; pp 3280--3291\relax
\mciteBstWouldAddEndPuncttrue
\mciteSetBstMidEndSepPunct{\mcitedefaultmidpunct}
{\mcitedefaultendpunct}{\mcitedefaultseppunct}\relax
\EndOfBibitem
\bibitem[Bach \latin{et~al.}(2019)Bach, Rodriguez, Liu, Luo, Shao, Xia, Sen, Ratner, Hancock, Alborzi, \latin{et~al.} others]{bach2019snorkel}
Bach,~S.~H.; Rodriguez,~D.; Liu,~Y.; Luo,~C.; Shao,~H.; Xia,~C.; Sen,~S.; Ratner,~A.; Hancock,~B.; Alborzi,~H.; others Snorkel drybell: A case study in deploying weak supervision at industrial scale. Proceedings of the 2019 International Conference on Management of Data. 2019; pp 362--375\relax
\mciteBstWouldAddEndPuncttrue
\mciteSetBstMidEndSepPunct{\mcitedefaultmidpunct}
{\mcitedefaultendpunct}{\mcitedefaultseppunct}\relax
\EndOfBibitem
\bibitem[Evensen \latin{et~al.}(2020)Evensen, Ge, and Demiralp]{evensen2020ruler}
Evensen,~S.; Ge,~C.; Demiralp,~C. Ruler: Data programming by demonstration for document labeling. Findings of the Association for Computational Linguistics: EMNLP 2020. 2020; pp 1996--2005\relax
\mciteBstWouldAddEndPuncttrue
\mciteSetBstMidEndSepPunct{\mcitedefaultmidpunct}
{\mcitedefaultendpunct}{\mcitedefaultseppunct}\relax
\EndOfBibitem
\bibitem[Li \latin{et~al.}(2021)Li, Ding, Shang, McAuley, and Feng]{li2021weakly}
Li,~J.; Ding,~H.; Shang,~J.; McAuley,~J.; Feng,~Z. Weakly supervised named entity tagging with learnable logical rules. \emph{arXiv preprint arXiv:2107.02282} \textbf{2021}, \relax
\mciteBstWouldAddEndPunctfalse
\mciteSetBstMidEndSepPunct{\mcitedefaultmidpunct}
{}{\mcitedefaultseppunct}\relax
\EndOfBibitem
\bibitem[Gao \latin{et~al.}(2022)Gao, Goswami, Chen, and Dubrawski]{gao2022classifying}
Gao,~C.; Goswami,~M.; Chen,~J.; Dubrawski,~A. Classifying unstructured clinical notes via automatic weak supervision. \emph{arXiv preprint arXiv:2206.12088} \textbf{2022}, \relax
\mciteBstWouldAddEndPunctfalse
\mciteSetBstMidEndSepPunct{\mcitedefaultmidpunct}
{}{\mcitedefaultseppunct}\relax
\EndOfBibitem
\bibitem[Varma and R{\'e}(2018)Varma, and R{\'e}]{varma2018snuba}
Varma,~P.; R{\'e},~C. Snuba: Automating weak supervision to label training data. Proceedings of the VLDB Endowment. International Conference on Very Large Data Bases. 2018; p 223\relax
\mciteBstWouldAddEndPuncttrue
\mciteSetBstMidEndSepPunct{\mcitedefaultmidpunct}
{\mcitedefaultendpunct}{\mcitedefaultseppunct}\relax
\EndOfBibitem
\bibitem[Das \latin{et~al.}(2020)Das, Chaba, Wu, Gandhi, Chau, and Chu]{das2020goggles}
Das,~N.; Chaba,~S.; Wu,~R.; Gandhi,~S.; Chau,~D.~H.; Chu,~X. Goggles: Automatic image labeling with affinity coding. Proceedings of the 2020 ACM SIGMOD International Conference on Management of Data. 2020; pp 1717--1732\relax
\mciteBstWouldAddEndPuncttrue
\mciteSetBstMidEndSepPunct{\mcitedefaultmidpunct}
{\mcitedefaultendpunct}{\mcitedefaultseppunct}\relax
\EndOfBibitem
\bibitem[Zhao \latin{et~al.}(2021)Zhao, Ding, and Feng]{inproceedings}
Zhao,~X.; Ding,~H.; Feng,~Z. GLaRA: Graph-based Labeling Rule Augmentation for Weakly Supervised Named Entity Recognition. 2021; pp 3636--3649\relax
\mciteBstWouldAddEndPuncttrue
\mciteSetBstMidEndSepPunct{\mcitedefaultmidpunct}
{\mcitedefaultendpunct}{\mcitedefaultseppunct}\relax
\EndOfBibitem
\bibitem[Boecking \latin{et~al.}(2021)Boecking, Neiswanger, Xing, and Dubrawski]{boecking2021interactive}
Boecking,~B.; Neiswanger,~W.; Xing,~E.; Dubrawski,~A. Interactive Weak Supervision: Learning Useful Heuristics for Data Labeling. International Conference on Learning Representations. 2021\relax
\mciteBstWouldAddEndPuncttrue
\mciteSetBstMidEndSepPunct{\mcitedefaultmidpunct}
{\mcitedefaultendpunct}{\mcitedefaultseppunct}\relax
\EndOfBibitem
\bibitem[Roberts \latin{et~al.}(2022)Roberts, Li, Huang, Adila, Schoenberg, Liu, Pick, Ma, Albarghouthi, and Sala]{roberts2022autowsbench}
Roberts,~N.; Li,~X.; Huang,~T.-H.; Adila,~D.; Schoenberg,~S.; Liu,~C.-Y.; Pick,~L.; Ma,~H.; Albarghouthi,~A.; Sala,~F. Auto{WS}-Bench-101: Benchmarking Automated Weak Supervision with 100 Labels. Thirty-sixth Conference on Neural Information Processing Systems Datasets and Benchmarks Track. 2022\relax
\mciteBstWouldAddEndPuncttrue
\mciteSetBstMidEndSepPunct{\mcitedefaultmidpunct}
{\mcitedefaultendpunct}{\mcitedefaultseppunct}\relax
\EndOfBibitem
\bibitem[Smith \latin{et~al.}(2022)Smith, Fries, Hancock, and Bach]{smith2022language}
Smith,~R.; Fries,~J.~A.; Hancock,~B.; Bach,~S.~H. Language Models in the Loop: Incorporating Prompting into Weak Supervision. \emph{arXiv preprint arXiv:2205.02318} \textbf{2022}, \relax
\mciteBstWouldAddEndPunctfalse
\mciteSetBstMidEndSepPunct{\mcitedefaultmidpunct}
{}{\mcitedefaultseppunct}\relax
\EndOfBibitem
\bibitem[Wang \latin{et~al.}(2021)Wang, Wang, Joty, and Hoi]{wang2021codet5}
Wang,~Y.; Wang,~W.; Joty,~S.; Hoi,~S.~C. Codet5: Identifier-aware unified pre-trained encoder-decoder models for code understanding and generation. \emph{arXiv preprint arXiv:2109.00859} \textbf{2021}, \relax
\mciteBstWouldAddEndPunctfalse
\mciteSetBstMidEndSepPunct{\mcitedefaultmidpunct}
{}{\mcitedefaultseppunct}\relax
\EndOfBibitem
\bibitem[Chen \latin{et~al.}(2021)Chen, Tworek, Jun, Yuan, Pinto, Kaplan, Edwards, Burda, Joseph, Brockman, \latin{et~al.} others]{chen2021evaluating}
Chen,~M.; Tworek,~J.; Jun,~H.; Yuan,~Q.; Pinto,~H. P. d.~O.; Kaplan,~J.; Edwards,~H.; Burda,~Y.; Joseph,~N.; Brockman,~G.; others Evaluating large language models trained on code. \emph{arXiv preprint arXiv:2107.03374} \textbf{2021}, \relax
\mciteBstWouldAddEndPunctfalse
\mciteSetBstMidEndSepPunct{\mcitedefaultmidpunct}
{}{\mcitedefaultseppunct}\relax
\EndOfBibitem
\bibitem[Nijkamp \latin{et~al.}(2022)Nijkamp, Pang, Hayashi, Tu, Wang, Zhou, Savarese, and Xiong]{Nijkamp2022CG}
Nijkamp,~E.; Pang,~B.; Hayashi,~H.; Tu,~L.; Wang,~H.; Zhou,~Y.; Savarese,~S.; Xiong,~C. CodeGen: An Open Large Language Model for Code with Multi-Turn Program Synthesis. \emph{arXiv preprint} \textbf{2022}, \relax
\mciteBstWouldAddEndPunctfalse
\mciteSetBstMidEndSepPunct{\mcitedefaultmidpunct}
{}{\mcitedefaultseppunct}\relax
\EndOfBibitem
\bibitem[Brown \latin{et~al.}(2020)Brown, Mann, Ryder, Subbiah, Kaplan, Dhariwal, Neelakantan, Shyam, Sastry, Askell, \latin{et~al.} others]{brown2020language}
Brown,~T.; Mann,~B.; Ryder,~N.; Subbiah,~M.; Kaplan,~J.~D.; Dhariwal,~P.; Neelakantan,~A.; Shyam,~P.; Sastry,~G.; Askell,~A.; others Language models are few-shot learners. \emph{Advances in neural information processing systems} \textbf{2020}, \emph{33}, 1877--1901\relax
\mciteBstWouldAddEndPuncttrue
\mciteSetBstMidEndSepPunct{\mcitedefaultmidpunct}
{\mcitedefaultendpunct}{\mcitedefaultseppunct}\relax
\EndOfBibitem
\bibitem[Zhang \latin{et~al.}(2021)Zhang, Yu, Li, Wang, Yang, Yang, and Ratner]{zhang2021wrench}
Zhang,~J.; Yu,~Y.; Li,~Y.; Wang,~Y.; Yang,~Y.; Yang,~M.; Ratner,~A. Wrench: A comprehensive benchmark for weak supervision. \emph{arXiv preprint arXiv:2109.11377} \textbf{2021}, \relax
\mciteBstWouldAddEndPunctfalse
\mciteSetBstMidEndSepPunct{\mcitedefaultmidpunct}
{}{\mcitedefaultseppunct}\relax
\EndOfBibitem
\bibitem[Vishwakarma \latin{et~al.}(2022)Vishwakarma, Roberts, and Sala]{vishwakarma2022Lifting}
Vishwakarma,~H.; Roberts,~N.; Sala,~F. Lifting Weak Supervision To Structured Prediction. \emph{arXiv preprint:2211.13375} \textbf{2022}, \relax
\mciteBstWouldAddEndPunctfalse
\mciteSetBstMidEndSepPunct{\mcitedefaultmidpunct}
{}{\mcitedefaultseppunct}\relax
\EndOfBibitem
\bibitem[Liu \latin{et~al.}(2023)Liu, Yuan, Fu, Jiang, Hayashi, and Neubig]{liu2023PromptingSurvey}
Liu,~P.; Yuan,~W.; Fu,~J.; Jiang,~Z.; Hayashi,~H.; Neubig,~G. Pre-Train, Prompt, and Predict: A Systematic Survey of Prompting Methods in Natural Language Processing. \emph{ACM Comput. Surv.} \textbf{2023}, \emph{55}\relax
\mciteBstWouldAddEndPuncttrue
\mciteSetBstMidEndSepPunct{\mcitedefaultmidpunct}
{\mcitedefaultendpunct}{\mcitedefaultseppunct}\relax
\EndOfBibitem
\bibitem[Dong \latin{et~al.}(2023)Dong, Li, Dai, Zheng, Wu, Chang, Sun, Xu, Li, and Sui]{ICL2023Survey}
Dong,~Q.; Li,~L.; Dai,~D.; Zheng,~C.; Wu,~Z.; Chang,~B.; Sun,~X.; Xu,~J.; Li,~L.; Sui,~Z. A Survey on In-context Learning. \textbf{2023}, \relax
\mciteBstWouldAddEndPunctfalse
\mciteSetBstMidEndSepPunct{\mcitedefaultmidpunct}
{}{\mcitedefaultseppunct}\relax
\EndOfBibitem
\bibitem[Arora \latin{et~al.}(2023)Arora, Narayan, Chen, Orr, Guha, Bhatia, Chami, and Re]{arora2023ask}
Arora,~S.; Narayan,~A.; Chen,~M.~F.; Orr,~L.; Guha,~N.; Bhatia,~K.; Chami,~I.; Re,~C. Ask Me Anything: A simple strategy for prompting language models. International Conference on Learning Representations. 2023\relax
\mciteBstWouldAddEndPuncttrue
\mciteSetBstMidEndSepPunct{\mcitedefaultmidpunct}
{\mcitedefaultendpunct}{\mcitedefaultseppunct}\relax
\EndOfBibitem
\bibitem[Wei \latin{et~al.}(2022)Wei, Wang, Schuurmans, Bosma, Ichter, Xia, Chi, Le, and Zhou]{wei2022chain-of-thought}
Wei,~J.; Wang,~X.; Schuurmans,~D.; Bosma,~M.; Ichter,~B.; Xia,~F.; Chi,~E.; Le,~Q.; Zhou,~D. Chain-of-Thought Prompting Elicits Reasoning in Large Language Models. \textbf{2022}, \relax
\mciteBstWouldAddEndPunctfalse
\mciteSetBstMidEndSepPunct{\mcitedefaultmidpunct}
{}{\mcitedefaultseppunct}\relax
\EndOfBibitem
\bibitem[Zhou \latin{et~al.}(2022)Zhou, Muresanu, Han, Paster, Pitis, Chan, and Ba]{zhou2022-auto-prompt}
Zhou,~Y.; Muresanu,~A.~I.; Han,~Z.; Paster,~K.; Pitis,~S.; Chan,~H.; Ba,~J. Large Language Models Are Human-Level Prompt Engineers. \textbf{2022}, \relax
\mciteBstWouldAddEndPunctfalse
\mciteSetBstMidEndSepPunct{\mcitedefaultmidpunct}
{}{\mcitedefaultseppunct}\relax
\EndOfBibitem
\bibitem[Sarsa \latin{et~al.}(2022)Sarsa, Denny, Hellas, and Leinonen]{sami2022-prompting-code}
Sarsa,~S.; Denny,~P.; Hellas,~A.; Leinonen,~J. Automatic Generation of Programming Exercises and Code Explanations Using Large Language Models. Proceedings of the 2022 ACM Conference on International Computing Education Research - Volume 1. New York, NY, USA, 2022; p 27–43\relax
\mciteBstWouldAddEndPuncttrue
\mciteSetBstMidEndSepPunct{\mcitedefaultmidpunct}
{\mcitedefaultendpunct}{\mcitedefaultseppunct}\relax
\EndOfBibitem
\bibitem[Denny \latin{et~al.}(2022)Denny, Kumar, and Giacaman]{paul2022prompting-code}
Denny,~P.; Kumar,~V.; Giacaman,~N. Conversing with Copilot: Exploring Prompt Engineering for Solving CS1 Problems Using Natural Language. \textbf{2022}, \relax
\mciteBstWouldAddEndPunctfalse
\mciteSetBstMidEndSepPunct{\mcitedefaultmidpunct}
{}{\mcitedefaultseppunct}\relax
\EndOfBibitem
\bibitem[Wang \latin{et~al.}(2021)Wang, Liu, Xu, Zhu, and Zeng]{wang-etal-2021-want-reduce}
Wang,~S.; Liu,~Y.; Xu,~Y.; Zhu,~C.; Zeng,~M. Want To Reduce Labeling Cost? {GPT}-3 Can Help. Findings of the Association for Computational Linguistics: EMNLP 2021. Punta Cana, Dominican Republic, 2021; pp 4195--4205\relax
\mciteBstWouldAddEndPuncttrue
\mciteSetBstMidEndSepPunct{\mcitedefaultmidpunct}
{\mcitedefaultendpunct}{\mcitedefaultseppunct}\relax
\EndOfBibitem
\bibitem[Xu \latin{et~al.}(2022)Xu, Alon, Neubig, and Hellendoorn]{xu2022systematic}
Xu,~F.~F.; Alon,~U.; Neubig,~G.; Hellendoorn,~V.~J. A systematic evaluation of large language models of code. Proceedings of the 6th ACM SIGPLAN International Symposium on Machine Programming. 2022; pp 1--10\relax
\mciteBstWouldAddEndPuncttrue
\mciteSetBstMidEndSepPunct{\mcitedefaultmidpunct}
{\mcitedefaultendpunct}{\mcitedefaultseppunct}\relax
\EndOfBibitem
\bibitem[Alberto \latin{et~al.}(2015)Alberto, Lochter, and Almeida]{alberto2015tubespam}
Alberto,~T.~C.; Lochter,~J.~V.; Almeida,~T.~A. Tubespam: Comment spam filtering on youtube. 2015 IEEE 14th international conference on machine learning and applications (ICMLA). 2015; pp 138--143\relax
\mciteBstWouldAddEndPuncttrue
\mciteSetBstMidEndSepPunct{\mcitedefaultmidpunct}
{\mcitedefaultendpunct}{\mcitedefaultseppunct}\relax
\EndOfBibitem
\bibitem[Ren \latin{et~al.}(2020)Ren, Li, Su, Kartchner, Mitchell, and Zhang]{ren2020denoising}
Ren,~W.; Li,~Y.; Su,~H.; Kartchner,~D.; Mitchell,~C.; Zhang,~C. Denoising multi-source weak supervision for neural text classification. \emph{arXiv preprint arXiv:2010.04582} \textbf{2020}, \relax
\mciteBstWouldAddEndPunctfalse
\mciteSetBstMidEndSepPunct{\mcitedefaultmidpunct}
{}{\mcitedefaultseppunct}\relax
\EndOfBibitem
\bibitem[Almeida \latin{et~al.}(2011)Almeida, Hidalgo, and Yamakami]{almeida2011contributions}
Almeida,~T.~A.; Hidalgo,~J. M.~G.; Yamakami,~A. Contributions to the study of SMS spam filtering: new collection and results. Proceedings of the 11th ACM symposium on Document engineering. 2011; pp 259--262\relax
\mciteBstWouldAddEndPuncttrue
\mciteSetBstMidEndSepPunct{\mcitedefaultmidpunct}
{\mcitedefaultendpunct}{\mcitedefaultseppunct}\relax
\EndOfBibitem
\bibitem[Dawid and Skene(1979)Dawid, and Skene]{Dawid:Skene:79}
Dawid,~A.~P.; Skene,~A.~M. Maximum Likelihood Estimation of Observer Error-Rates Using the EM Algorithm. \emph{Applied Statistics} \textbf{1979}, \emph{28}, 20--28\relax
\mciteBstWouldAddEndPuncttrue
\mciteSetBstMidEndSepPunct{\mcitedefaultmidpunct}
{\mcitedefaultendpunct}{\mcitedefaultseppunct}\relax
\EndOfBibitem
\bibitem[Fu \latin{et~al.}(2020)Fu, Chen, Sala, Hooper, Fatahalian, and R\'{e}]{threerius}
Fu,~D.~Y.; Chen,~M.~F.; Sala,~F.; Hooper,~S.~M.; Fatahalian,~K.; R\'{e},~C. Fast and Three-Rious: Speeding up Weak Supervision with Triplet Methods. Proceedings of the 37th International Conference on Machine Learning. 2020\relax
\mciteBstWouldAddEndPuncttrue
\mciteSetBstMidEndSepPunct{\mcitedefaultmidpunct}
{\mcitedefaultendpunct}{\mcitedefaultseppunct}\relax
\EndOfBibitem
\end{mcitethebibliography}
\bibliographystyle{achemso}

\end{document}